\documentstyle{article}
\begin{document}
\title{Is Word Sense Disambiguation just one more NLP task?}
\author{Yorick Wilks}
\date{}
\maketitle

\begin{abstract}
The paper compares the tasks of part-of-speech (POS) tagging and word-sense-tagging or disambiguation (WSD), and argues that the tasks 
are not related by fineness of grain or anything like that, but are 
quite different kinds of task, particularly because there is nothing 
in POS corresponding to sense novelty.  The paper also argues for the
reintegration of sub-tasks that are being separated for evaluation.
\end{abstract}

\section{Introduction}
{\parindent 0pt

I want to make clear right away that I am not writing as a sceptic
about word-sense disambiguation (WSD) let alone as a recent convert: on
the contrary, since my PhD thesis was on the topic thirty years ago.
That (Wilks, 1968) was what we would now call a classic AI toy system
approach,  one that used techniques later called Preference Semantics,
but applied to real newspaper texts, as controls on the philosophical
texts that were my real interest at the time. But it did attach single
sense representations to words drawn from a polysemous lexicon of 800
or so. If Boguraev was right, in his informal survey twelve years ago,
that the  average NLP lexicon was under fifty words, then that work was
ahead of its time and I do therefore  have a longer commitment to, and
perspective on, the topic than most, for whatever that may be worth!.
}

I want to raise some general questions about WSD as a task, aside from all the
busy work in SENSEVAL: questions that should make us worried and wary
about what we are doing here, but definitely NOT stop doing it. I can
start by reminding us all of the obvious ways in which WSD is not like
part-of-speech (POS) tagging, even though the two tasks are plainly connected
in information terms, as Stevenson and I pointed out in (Wilks and
Stevenson, 1998a), and were widely misunderstood for doing so. From
these differences, of POS and WSD, I will conclude that WSD is not just
one more partial task to be hacked off the body of NLP and solved.
What follows acknowledges that Resnik and Yarowsky made a similar
comparison in 1997 (Resnik and Yarowsky, 1997) though this list is a
little different from theirs:

\begin{enumerate}
\item
There is broad agreement about POS tags in that, even for those
committed to differing sets, there is little or no dispute that they
can be put into one-many correspondence. That is not generally accepted
for the sets of senses for the same words from different lexicons.

\item
There is little dispute that humans can POS tag to a high degree of
consistency, but again this is not universally agreed for WS tagging,
as various email discussions leading up to this workshop have shown.
I'll come back to this issue below, but its importance cannot
be exaggerated --- if humans cannot do it then we are wasting our time
trying to automate it. I
assume that fact is clear to everyone: whatever maybe the case in
robotics or fast arithmetic, in the NL parts of AI there is no point
modelling or training for skills that humans do not have!

\item
I do not know the genesis of the phrase \lq\lq lexical tuning," but the
phenomenon has been remarked, and worked on, for thirty years and
everyone seems agreed that it happens, in the sense that human
generators create, and human analysers understand, words in quite new
senses, ungenerated before or, at least, not contained in the
point-of-reference lexicon, whether that be thought of as in the head
or in the computer.  Only this view is consistent with the evident
expansion of sense lists in dictionaries with time; these new additions
cannot plausibly be taken as established usages not noticed before. 
\end{enumerate}

{\parindent 0pt
If this is the case, it seems to mark an absolute difference from POS
tagging (where novelty does not occur in the same way), and that should
radically alter our view of what we are doing here, because we cannot
apply the standard empirical modelling method to that kind of novelty.
}

The now standard empirical paradigm of [mark-up, model/train, and test]
assumes prior markup, in the sense of a positive answer to the question
(2) above. But we cannot, by definition, mark up for new senses, those
not in the list we were initially given, because the text analysed
creates them, or they were left out of the source from which the mark
up list came.  If this phenomenon is real, and I assume it is, it sets
a limit on phenomenon (2), the human ability to pre-tag with senses,
and therefore sets an upper bound on the percentage results we can
expect from WSD, a fact that marks WSD out quite clearly from POS
tagging.

The contrast here is in fact quite subtle as can be seen from the
interesting intermediate case of semantic tagging: which is the task 
of attaching semantic,
rather than POS, tags to words automatically, a task which can then be
used to do more of the WSD task (as in Dini et al., 1998) than POS
tagging can, since the ANIMAL or BIRD  versus MACHINE tags can then
separate the main senses of \lq\lq crane". In this case, as with POS, one
need not assume novelty in the tag set, but must allow for novel
assignments from it to corpus words e.g. when a word like \lq\lq dog" or
\lq\lq pig" was first used in a human sense. It is just this sense of novelty
that POS tagging does also have, of course, since a POS tag like VERB can be
applied to what was once only a noun, as with \lq\lq ticket". This kind of
novelty, in POS and semantic tagging, can be pre-marked up with a fixed
tag inventory, hence both these techniques differ from genuine sense
novelty which cannot be premarked.
 
As I said earlier, the thrust of these remarks is not intended
sceptically, either about WSD in particular, or about the empirical
linguistic agenda of the last ten years more generally.  I assume the
latter has done a great deal of good to NLP/CL: it has freed us from
toy systems and fatuous example mongering, and shown that more could be
done with superficial knowledge-free methods than the whole AI
knowledge-based-NLP tradition ever conceded: the tradition in which
every example, every sentence, had in principle to be subjected to the
deepest methods. Minsky and McCarthy always argued for that, but it
seemed to some even then an implausible route for any
least-effort-driven theory of evolution to have taken. The caveman
would have stood paralysed in the path of the dinosaur as he downloaded
deeper analysis modules, trying to disprove he was only having a
nightmare.

However, with that said, it may be time for some corrective: time to
ask not only how we can continue to slice off more fragments of partial NLP
as tasks to model and evaluate, but also how to reintegrate them  for
real tasks that humans undoubtedly can evaluate reliably, like MT and
IE, and which are therefore unlike some of the partial tasks we have
grown used to (like syntactic parsing) but on which normal language
users have no views at all, for they are expert-created tasks, of
dubious significance outside a wider framework. It is easy to forget
this because it is easier to keep busy, always moving on. But there are
few places left to go after WSD:--empirical pragmatics has surely
started but may turn out to be the final leg of the journey.
 
Given the successes of empirical NLP at such a wide range of tasks, it
is not to soon to ask what it is all for, and to remember that, just
because machine translation (MT) researchers complained long ago that
WSD was one of their main problems, it does not follow that high level
percentage success at WSD will advance MT. It may do so, and it is
worth a try, but we should remember that Martin Kay warned years ago
that no set of individual solutions to computational semantics, syntax,
morphology etc. would necessarily advance MT. However, unless we put
more thought into reintegrating the new techniques developed in the
last decade we shall never find out.

\section{Can humans sense tag?}

{\parindent 0pt
I wish now to return to two of the topics raised above: first, the
human task: itself.
}

It seems obvious to me that, aside from the problems of tuning and
other phenomena that go under names like vagueness, humans, after
training, can sense-tag texts at reasonably high levels and reasonable
inter-annotator consistency. They can do this with alternative sets of
senses for words for the same text, although it may be a task where
some degree of training and prior literacy are essential, since some
senses in such a list are usually not widely known to the public. This 
should not be
shocking: teams of lexicographers in major publishing houses constitute
literate, trained teams and they can normally achieve agreement
sufficient for a large printed dictionary for publication (about sense
sets, that is, a closely related skill to sense-tagging).  Those averse
to claims about training and expertise here should remember that most
native speakers cannot POS tag either, though there seems substantial
and uncontentious consistency among the trained.

There is strong evidence for this position on tagging ability, which
includes (Green, 1989 see also Jorgensen, 1990) and indeed the
high figures obtained for small word sets by the techniques pioneered
by Yarowsky (Yarowsky, 1995). Many of those figures rest on forms of
annotation (e.g. assignment of words to thesaurus head sets in Roget),
and the general plausibility of the methodology  serves to confirm the
reality of human annotation (as a consistent task) as a side effect.

The counterarguments to this have come explicitly from the writings of
Kilgarriff (1993), and sometimes implicitly from the work of those who
argue from the primacy of lexical rules or of notions like vagueness in
regard to WSD. In Kilgarriff's case I have argued elsewhere
(Wilks, 1997) that the figures he produced on human annotation are
actually consistent with very high levels of human ability to sense-tag
and are not counter-arguments at all, even though he seems to remain
sceptical about the task in his papers. He showed only that for most
words there are some contexts for which humans cannot assign a sense,
which is of course not an argument against the human skill being
generally successful.

On a personal note, I would hope very much to be clearer when I see his
published reaction to the SENSEVAL workshop what his attitude to WSD
really is. In writing he is a widely published sceptic, in the flesh he
is the prime organiser of this excellent event (SENSEVAL Workshop)
to test a skill he may,
or may not, believe in.  There need be no contradiction there, but a
fascinating question about motive lingers in the air. Has he set all
this up so that WSD can destroy itself when rigourously tested? One
does not have to be a student of double-blind tests, and the role of
intention in experimental design, to take these questions seriously,
particularly as he has designed the SENSEVAL methodology and the use 
of the data
himself. The motive question here is not mere ad hominem argument but a
serious question needing an answer.

These are not idle questions, in my view, but go to the heart of what
the SENSEVAL workshop is for: is it to show how to do better at WSD, or
is to say something about wordsense itself (which might involve saying
that you cannot do WSD by computer at all, or cannot do it well enough
to be of interest?).

In all this discussion we should remember that, if we take the
improvement of (assessable) real tasks as paramount, those like MT,
Information Retrieval and Information Extraction (IE), then it may not
in the end matter whether humans are ever shown psycholinguistically to
need POS tagging or WSD for their own language performance;--there is
much evidence they do not. But that issue is wholly separate from what
concerns us here; it may still be useful to advance MT/IE via partial
tasks like WSD, if they can be shown performable, assessable, and
modelable by computers, no matter how humans turn out to work.

The implicit critique of the broadly positive position above (i.e. that
WSD can be done by people and machines and we should keep at it)
sometimes seems to come as well from those who argue (a) for the
inadequacy of lexical sense sets over productive lexical rules and (b)
for the inherently vague quality of the difference between senses of a
given word. I believe both these approaches are muddled if their
proponents conclude that WSD is therefore fatally flawed as a task;- and
clearly not all do since some of them are represented here as
participants.

\section{Lexical Rules}

Lexical rules go back at least to Givon's (1967) thirty-year old
sense-extension rules and they are in no way incompatible with a
sense-set approach, like that found in a classic dictionary. Such sense
sets are normally structured (often by part of speech and by general
and specific senses) and the rules are, in some sense, no more than a
compression device for predicting that structuring. But the set
produced by any set of lexical rules is still a set, just as a
dictionary list of senses is a set, albeit structured. It is mere
confusion to think one is a set and one not: Nirenburg and Raskin
(1997) have pointed out that those who argue against lists of senses
(in favour of rules, e.g. Pustejovsky 1995) still produce and use such
lists. What else could they do?

I myself cannot get sufficient clarity at all on what the lexical 
rule approach,
whatever its faults or virtues, has to do with WSD? The email
discussion preceding this workshop showed there were people who think
the issues are connected, but I cannot see it, but
would like to be better informed before I go home from here. If their
case is that rules can predict or generate new senses then their
position is no different (with regard to WSD) from that of anyone else
who thinks new senses important, however modelled or described. The
rule/compression issue itself has nothing essential to do with WSD: it
is simply one variant of the novelty/tuning/new-sense/metonymy problem,
however that is described.

The vagueness issue is again an old observation, one that, if taken
seriously, must surely result in a statistical or fuzzy-logic approach
to sense discrimination, since only probabilistic (or at least
quantitative) methods can capture real vagueness. That, surely, is the
point of the Sorites paradox:  there can be no plausible or rational
qualitatively-based criterion (which would include any quantitative
system with clear limits: e.g. tall = over 6 feet) for demarcating
\lq\lq tall", \lq\lq green" or any inherently vague concept.

If, however, sense sets/lists/inventories are to continue to play a
role, vagueness can mean no more than highlighting what all systems of
WSD must have, namely some parameter or threshold for the assignment to
one of a list of senses versus another, or setting up a new sense in
the list.  Talk of vagueness adds nothing specific to help that process
for those who want to assign on some quantitative basis to one sense
rather than another; algorithms will capture the usual issue of 
tuning to see what works and fits our intuitions.
 
Vagueness would be a serious concept only if the whole sense list for a
word (in rule form or not) was abandoned in favour of
statistically-based unsupervised clusters of usages or contexts. 
There have been
just such approaches to WSD in recent years (e.g. Bruce and Wiebe,
1994, Pedersen and Bruce, 1997, Schuetze \& Pederson, 1995)  and the
essence of the idea goes back to Sparck Jones 1964/1986) but such an
approach would find it impossible to take part in any competition like
SENSEVAL because it would inevitably deal in nameless entities which
cannot be marked up for.
 
Vague and Lexical Rule based approaches also have the consequence that
all lexicographic practice is, in some sense,  misguided: dictionaries
according to such theories are fraudulent documents that could not help 
users,
whom they systematically mislead by listing senses. Fortunately, the
market decides this issue, and it is a false claim. Vagueness in WSD is
either false (the last position) or trivial, and known and utilised
within all methodologies.

This issue owes something to the systematic ignorance of its own
history so often noted in AI.  A discussion email preceding this
workshop referred to the purported benefits of underspecification in
lexical entries, and how recent formalisms had made that possible. How
could  anyone write such a thing in ignorance of the 1970s and 80s work
on incremental semantic interpretation of Hirst, Mellish and Small
(Hirst, 1987; Mellish, 1983; Small et al., 1988) among others?
  
None of this is a surprise to those with AI memories more than a few
weeks long: in our field people read little outside their own
notational clique, and constantly \lq\lq rediscover" old work with a new
notation. This leads me to my final point which has to do, as I noted
above, with the need for a fresh look at technique integration for real
tasks. We all pay lip service to this while we spend years on
fragmentary activity, arguing that that is the method of science. Well,
yes and no, and anyway WSD is not science: what we are doing is
engineering and the scientific method does not generally work there, since
engineering is essentially integrative, not analytical. We often write
or read of \lq\lq hybrid" systems in NLP, which is certainly an integrative
notion, but we have little clear idea of what it means. If statistical
or knowledge-free methods are to solve some or most cases of any
linguistic phenomenon, like WSD, how  do we then locate that subclass
of the phenomena that other, deeper, techniques like AI and
knowledge-based reasoning are then to deal with? Conversely, how can
we know which cases the deeper techniques cannot or need not deal with?
If there is an upper bound to empirical methods, and I have argued that
that will be lower for WSD than for some other NLP tasks for the reasons 
set out above, then how can we pull in other techniques smoothly and
seamlessly for the \lq\lq hard" examples?

The experience of POS tagging, to return to where we started, suggests
that rule-driven taggers can do as well as purely machine 
learning-based taggers,
which, if true, suggests that symbolic methods, in a broad sense, might
still be the right approach for the whole task. Are we yet sure this is
not the case for WSD? I simply raise the question. Ten years ago, it
was taken for granted in most of the AI/NLP community that
knowledge-based methods were essential for serious NLP. Some of the
successes of the empirical program (and especially the TIPSTER program)
have caused many to reevaluate that assumption. But where are we now,
if a real ceiling to such methods is already in sight? 
Information Retrieval languished
for years, and maybe still does, as a technique with a practical use but an
obvious ceiling, and no way of breaking through it; there was really
nowhere for its researchers to go. But that is not quite true for us,
because the claims of AI/NLP to offer high quality at NLP tasks have
never been really tested. They have certainly not failed, just got left
behind in the rush towards what could be easily tested!

\section{Large or Small-scale WSD?}

Which brings me to my final point: general versus small-scale WSD. Our
group is one of the few that has insisted on continuing with general
WSD: the tagging and test of all content words in a text, a group that
includes CUP, XERC-Grenoble and CRL-NMSU. We currently claim about  90\%
correct sense assignment (Wilks and Stevenson, 1998b) and do not expect
to be able to improve much on that for the reasons set out above; we believe
the rest is AI or lexical tuning! The general argument for continuing
with the all-word paradigm, rather than the highly successful paradigm of
Yarowsky et al., is that that is the real task, and there is no firm
evidence that the small scale will scale up to the large because much of
sense-disambiguation is mutual between the words of the text, which
cannot be used by the small set approach. I am not sure this argument
is watertight but it seems plausible to me.

Logically, if you claim to do all the content words you ought, in principle,
to be able to enter a contest like SENSEVAL that does only some of the
words with an unmodified system. This is true, but you will also expect
to do worse, as you have not have had as much training data for the
chosen word set.  Moreover you will have to do far more preparation to
enter if you insist, as we would, on bringing the engines and data into
play for all the training and test set words; the effort is that much
greater and it makes such an entry self-penalising in terms of both
effort and likely outcome, which is why we decided not to enter in the
first round, regretfully, but just to mope and wail at the sidelines.
The methodology chosen for SENSEVAL was a natural reaction to the lack
of training and test data for the WSD task, as we all know, and that is
where I would personally like to see effort put in the future, so that
everyone can enter all the words; I assume that would be universally
agreed to if the data were there. It is a pity, surely, to base the
whole structure of a competition on the paucity of the data.

\section{Conclusion}

What we would like to suggest positively is that we cooperate to
produce more data, and use existing all-word systems, like Grenoble,
CUP, our own and others willing to join, possibly in combination, so as
to create large-scale tagged data quasi-automatically, rather in the
way that the Penn tree bank was produced with the aid of parsers,
not just people.

We have some concrete suggestions as to how this can be done, and done
consistently, using not only multiple WSD systems but also by cross
comparing the lexical resources available, e.g. WordNet (or
EuroWordNet) and a major monolingual dictionary. We developed our own
reasonably large test/training set with the WordNet-LDOCE sense
translation table (SENSUS, Knight and Luk, 1994) from ISI. Some sort of
organised effort along those lines, before the next SENSEVAL, would
enable us all to play on a field not only level, but much larger.

\end{document}